%% file: 1main.tex
\documentclass[conference]{IEEEtran}
\IEEEoverridecommandlockouts
\usepackage{cite}
\usepackage[numbers,sort&compress]{natbib}
\usepackage{amsmath,amssymb,amsfonts}
\usepackage{algorithmic}
\usepackage{amsthm}
\usepackage{enumerate}
\usepackage{amsfonts}
\usepackage{color}
\usepackage{hyperref}
\usepackage[justification=centering]{caption}
\usepackage[ruled,linesnumbered]{algorithm2e}
\usepackage{graphicx}
\usepackage{textcomp}
\usepackage{xcolor}
\usepackage{multicol}
\usepackage{amsthm}
\usepackage{caption}
\usepackage{subfigure}
\usepackage{tabularx}
\usepackage{cite}
\usepackage{url}
\usepackage{setspace}
\usepackage[utf8]{inputenc}
\usepackage{multirow}
\usepackage{tikz}
\usetikzlibrary{positioning}

\usepackage{booktabs}

\def\BibTeX{{\rm B\kern-.05em{\sc i\kern-.025em b}\kern-.08em
    T\kern-.1667em\lower.7ex\hbox{E}\kern-.125emX}}
\begin{document}



\title{A Multi-Source Multi-View Learning Model Framework for Airbnb Price Prediction}
\title{A Multi-Source Information Learning Framework for Airbnb Price Prediction}

\author{\IEEEauthorblockN{Lu Jiang$^{1}$, Yuanhan Li$^{1}$, Na Luo$^{1}$, Jianan Wang$^{2,*}$, Qiao Ning$^{3,*}$}
\IEEEauthorblockA{\textit{$^1$Information Science and Technology, Northeast Normal University, Changchun} \\
\textit{$^2$College of Physics, Northeast Normal University, Changchun} \\
\textit{$^3$Information Science and Technology, Dalian Maritime University, Dalian} \\
\textit{\{jiangl761, liyh447, luon110, wangjn\}@nenu.edu.cn, ningq669@dlmu.edu.cn}}
\textit{Corresponding author*}
}

\maketitle


\input{2abstract}
\input{3introduction}

\input{4preliminary}

\input{5method}

\input{6experiment}

\input{7relatedwork}

\input{8conclusion}

\bibliographystyle{IEEEtran}
\bibliography{ref}

\end{document}

%% file: 2abstract.tex
\begin{abstract}

With the development of technology and sharing economy, Airbnb as a famous short-term rental platform, has become the first choice for many young people to select.
The issue of Airbnb's pricing has always been a problem worth studying. 
While the previous studies achieve promising results, there are exists deficiencies to solve. 
Such as, (1) the feature attributes of rental are not rich enough; 
(2) the research on rental text information is not deep enough; 
(3) there are few studies on predicting the rental price combined with the point of interest(POI) around the house.
To address the above challenges, we proposes a multi-source information embedding(MSIE) model to predict the rental price of Airbnb. 
Specifically, we first selects the statistical feature to embed the original rental data. 
Secondly, we generates the word feature vector and emotional score combination of three different text information to form the text feature embedding.
Thirdly, we uses the points of interest(POI) around the rental house information generates a variety of spatial network graphs, and learns the embedding of the network to obtain the spatial feature embedding.
Finally, this paper combines the three modules into multi source rental representations, and uses the constructed fully connected neural network to predict the price. 
The analysis of the experimental results shows the effectiveness of our proposed model.

\end{abstract}

%% file: 3introduction.tex
\section{Introduction}

Accommodation sharing systems are being introduced to more and more cities recently, and therefore they have generated huge amounts of data.
Airbnb is an online marketplace for sharing home and experience which is suffering from the chaotic pricing problem. 
Tenants need to know the reasonable price of this rental house to prevent being deceived. 
The homeowner needs to customize a reasonable price for their short-term rental house to attract more customers.
Therefore, airbnb price prediction plays a key role in accommodation sharing systems.
However, rapid increase in the number of tenants and homeowners makes traditional manual-based methods~\cite{RePEc} time-consuming and inefficient.
Computational methods have received more attention for accurate airbnb price prediction~\cite{DBLP:journals/ijbidm/MoranoT13}. 

Computational methods for price prediction can be mainly divided into two categories:
(1) feature-based methods~\cite{DBLP:conf/dasfaa/XuHWFLCWY19},
and (2) deep learning methods~\cite{DBLP:journals/tdasci/XuFWWHFY20,DBLP:conf/icdm/FuGZYLXY14}.
In feature-based methods, various types of features extraction strategies are utilized to extract price correlated features for tenants and homeowners.
Feature-based methods transform price prediction into a machine learning methods, such as support vector machine(SVM) and random forest.
For instance, in order to distinguish from the traditional method of formulating prices, Li {\it et al.} selects rough set (RS) and SVM algorithms to establish a new mathematical model of pricing on the basis of hedonic price~\cite{2008yanqingli}.  
PR Kalehbasti{\it et al.} proposed a price prediction model using machine learning, deep learning, and natural language processing techniques to embed the features of the rentals, owner characteristics, and the customer reviews~\cite{Rezazadeh_Kalehbasti_2021}.
Deep learning methods, which use multi-layer neural network to map the correlation between input features and output results. 
For instance, Chen {\it et al.} applied auto regressive integrated moving average model to generate the baseline while LSTM networks to build prediction model~\cite{DBLP:journals/corr/abs-1709-08432}. 

However, the research of airbnb price prediction based on feature-based methods consider the single feature in most cases.
With the development of representation learning~\cite{DBLP:journals/tist/WangFZLL18,DBLP:conf/kdd/WangFXL19}, the spatial embedding~\cite{2020Incremental,DBLP:conf/kdd/WangFZWZA18} have received more attention.
There has been work to model the statistical features, text feature and spatial features related to housing prices, but there is no unified framework to integrate the above features.
Based on the above disadvantages, we proposes a prediction model based on multi-source information embedding to study the Airbnb price problem. 
The major contributions are summarized below.

\begin{itemize}
    \item Firstly, in order to obtain the best feature set, this paper selects the features of the house itself to obtain statistical information features.
    
    \item Secondly, the text information in this paper is divided into three categories, and the house description and landlord introduction are converted into feature matrix.
    The tenant reviews are then converted into sentiment scores about each house. 
    
    \item Then, we uses different types of point-of-interest (POI) data and houses to form various spatial network graphs and learns their network embeddings to obtain spatial information features.  
    
    \item Finally, we combines these three types of feature embeddings are combined into multi-resource housing features as input, and the neural network constructed in this paper is used for price prediction. The effectiveness of our model is demonstrated with two real data.
\end{itemize}

%% file: 4preliminary.tex
\section{Preliminary}
We first introduce some key definitions and the problem definition. 
Then, we present the overview of the proposed method.

\subsection{Definitions and Problem Statement}

\newtheorem{definition}{Definition}
\begin{definition}
{\bf Statistic Feature}
The statistics feature constructed by our model is $\bf S$ =$\left(s_{1}, s_{2}, ..., s_{n}\right)$, where $s_i$ is the preprocessed listing features includes 'host\_since', ‘host\_is\_superhost', 'verification', etc. 
\end{definition}

\begin{definition}

{\bf Text Feature}
There are three types of text features: listing description, host introduction, and tenant review. 
We convert listing description and host introduction into feature vector
$\bf L$ =$\left(l_{1}, l_{2}, ..., l_{n}\right)$ and 
$\bf H$ =$\left(h_{1}, h_{2}, ..., h_{n}\right)$, and transform the tenant review to sentiment score 
$\bf R$ =$\left(r_{1}, r_{2}, ..., r_{n}\right)$. 
Thus, we define the text features as $\bf T =\left ( \bf L,\bf H,\bf R \right )$. 
\end{definition}

\begin{definition}
{\bf Spatial Feature}
We first combine each rental house and the POI with in 1,000m around it into a spatial network $G = (V,E,W)$.
Then, we learn network embedding through SDNE~\cite{DBLP:conf/kdd/WangC016}, and get the spatial feature matrix $\bf P$ =$\left(p_{1}, p_{2}, ..., p_{n}\right)$.

\end{definition}

\begin{definition}
{\bf Problem Statement} 
In this paper, we study the problem of airbnb price prediction.
We formulate the problem as a multi-source feature embedding task. 
Formally, we aim to find a mapping function
$f: (S,T,P) \rightarrow V$ that takes the statistic feature $S$, text feature $T$, spatial feature $P$ as input, and outputs a unified vectorized representations $V$, for
predicting the specific listing price.

\end{definition}

\subsection{Framework Overview}

\begin{figure*}[!t]
	\centering
	\includegraphics[width=0.95\linewidth]{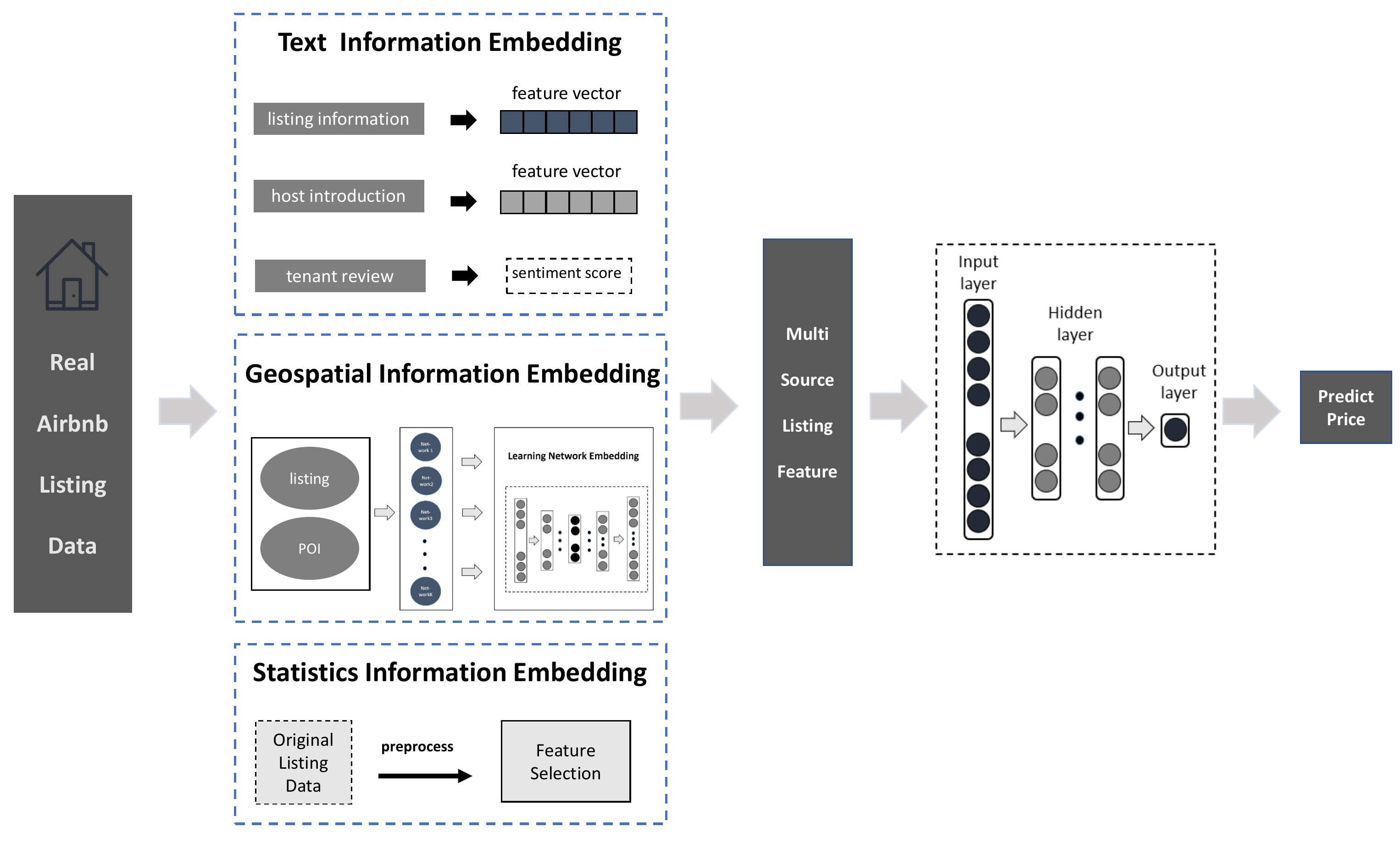}
	\captionsetup{justification=centering}
	\caption{Framework Overview.}
	\label{fig:framework overview}
\end{figure*}

Figure~\ref{fig:framework overview} shows an overall framework for the multi-source feature embedding.
Specifically, we embed the original data from three aspects.
(1) For the statistical feature embedding, we uses Lasso CV to select the feature set with the rental house feature.
(2) For the text feature embedding, we divides the text feature into three categories, include house description, landlord introduction and tenant comments. 
Through the negative sampling CBOW model, the house description and landlord introduction are converted into word feature vectors, and the Bayesian model based on naive Bayesian principle is used to convert tenant comments into emotional scores, we combine them as the text feature.
(3) For the spatial feature embedding, we collects different types of POIs, and combines the POI of each house and the surrounding area within 1,000m into a spatial network. Through the SDNE model to learn the spatial feature.
(4) Three different features are combined into a multi-source feature and input into the neural network to obtain the final rental price. 

%% file: 5method.tex
\section{Multi-Source Information Learning}

In this section, we introduce the core architecture of
our framework as follows:
(1) statistic feature embedding;
(2) text feature embedding;
(3) spatial feature embedding.

\subsection{Statistics Feature Embedding}
Each house's statistic feature is represented by a 245-
dimensional vector which describes the listing of a house, including listing\_{id}, host\_{id}, host\_{since}, host\_{response}\_{rate}, host\_{is}\_{superhost}, host\_{has}\_{profile}\_{pic}, host\_{identity}\_{verified}, bathrooms, bedrooms, latitude, longitude, accommodates, 		 security\_{deposit}, guests\_{included}, verification, etc.

We use the Lasso CV to do the feature set selection.
The loss function is defined as follows:

\begin{equation}
\text { obj }=\frac{1}{2} \sum_{i=1}^n\left(\boldsymbol{y}_{\mathbf{i}}-\boldsymbol{w}^T \boldsymbol{x}_{\mathbf{i}}\right)^2+\alpha \sum_{j=1}^m\left|w_{\mathrm{i}}\right|
\end{equation}

where $n$ is the number of houses, $m$ is the number of parameters, $\alpha$ is the regularization coefficient, $\alpha \sum_{j=1}^m\left|w_{\mathrm{i}}\right|$ is the L1 regularization term, $y_i$ is the rental price, $x_i$ is the statistical features of rental housing, $w$ is the coefficient matrix of rental housing features, $x_i = s_i$. 
Statistical feature matrix $\bf S$ =$\left(s_{1}, s_{2}, ..., s_{n}\right)$. 
Lasso CV can compress the coefficients of unimportant features to 0, realizing the purpose of feature selection, and ultimately leaving the important statistical feature set that this paper wants.

\subsection{Text Feature Embedding}
We extract three types of text data from the original data, there are listing description, host introduction and tenant review. 
Listing description mainly about introducing the location of the house, surrounding environment, indoor layout and housing regulations, etc. 
Host introduction mainly introduces the age, height, occupation, hobbies and personality of the host, and tenant review expresses the tenants’ feelings about housing rentals and the evaluation of the host’s attitudes. 
Since tenant review contain emotional value, we use two different methods to model text feature.
We first use CBOW~\cite{DBLP:conf/cikm/LuoXG14} model to embed the text feature of listing description, host introduction.
We selects the Wikipedia Chinese thesaurus after preprocessing as the training corpus $W$, the objective function is defined as follows:

\begin{equation}
\mathcal{L}=\sum_{c \in W}\left\{\log \left[\sigma\left(x_c^T \theta^c\right)\right]+\sum_{u \in N E G(c)} \log \left[\sigma\left(-x_c^T \theta^u\right)\right]\right\}
\end{equation}

Then the above objective function is optimized by using the random gradient rise method to obtain:

\begin{equation}
\mathcal{L}(c, u)=L^c(u) \log \left[\sigma\left(x_c^T \theta^u\right)\right]+\left[1-L^c(u)\right] \log \left[1-\sigma\left(x_c^T \theta^u\right)\right]
\end{equation}

Then calculate the gradient of $\mathcal{L}(c, u)$ to obtain:

\begin{equation}
\boldsymbol{v}(\tilde{c}):=\boldsymbol{v}(\tilde{c})+\eta \sum_{u \in\{c\} \cup N E G(c)} \frac{\partial \mathcal{L}(c, u)}{\partial x_c}
\end{equation}

In this paper, we set the dimension of the word vector as 100.
$l$ and $h$ represent the embedding of listing description, host introduction, respectively.

\begin{equation}
l=\frac{1}{Z} \sum_{i=1}^Z v\left(\tilde{c}_l\right)
\end{equation}

\begin{equation}
h=\frac{1}{z} \sum_{i=1}^z v\left(\tilde{c}_h\right)
\end{equation}

Therefore, we get the text information feature matrix of the listing description and host introduction: $\bf L$ =$\left(l_{1}, l_{2}, ..., l_{n}\right)$ and $\bf H$ =$\left(h_{1}, h_{2}, ..., h_{n}\right)$.

For the tenant review embedding, since it contains strong emotional expression, in order to reflect whether the tenants' evaluation of the house is positive or negative, we uses the  naive Bayes method to generate the corresponding emotional score $r \in [0,1]$ for each house, where 0 represents the negative and 1 represents the positive.
Specifically, the probability that a tenant review text belongs to the positive class can be expressed as:

\begin{equation}
P\left(p o s \mid c_1, \ldots c_d\right)=\frac{P\left(c_1, \ldots c_d \mid p o s\right) P(p o s)}{P\left(c_1, \ldots c_d\right)}
\end{equation}

After simplifying the above formula, we can obtain:

\begin{equation}
 P\left(p o s \mid c_1, \ldots c_d\right)=\frac{1}{1+\gamma}
\end{equation}

In this work, a text represents a tenant's review, and a tenant has many reviews, so the emotional score of a tenant can be expressed as:

\begin{equation}
r=\frac{1}{q} \sum_{i=1}^q P\left(\operatorname{pos} \mid c_1, \ldots c_d\right)
\end{equation}

where $q$ represents the number of reviews on a rental, $d$ represents the total number of words in a review, and $P\left(\operatorname{pos} \mid c_1, \ldots c_d\right)$ represents the probability that the review belongs to the category of positive emotions. 
Therefore, the emotional score vector of tenant reviews can be expressed as $\bf R$ =$\left(r_{1}, r_{2}, ..., r_{n}\right)$.

\subsection{Spatial Feature Embedding}

We proposes a method to learn spatial embedding. 
First, POI is divided into 8 different types, 
and the rented houses and the surrounding different type POIs form a spatial network. 
Then learn the network embedding of these spatial graphs through SDNE model. This method can accurately capture the spatial features related to important POIs such as scenic spots and railway stations.

We uses Euclidean distance to calculate the weight $W$ between house and poi as follows:

\begin{equation}
W=R \cdot \arccos (\text { dis }) \cdot \pi / 180
\end{equation}

where $d i s=\sin ( Lat_A) \sin ( Lat_B) \cos ( Lon_A- Lon_B)+\cos ( Lat_A) \cos ( Lat_B )$, the two types of nodes, A and B, represent the rental houses and POI respectively. 
$Lon_A $ and $Lon_B$ are their longitudes, $Lat_A$ and $Lat_B$ are their latitudes, and $R$ is the average radius of the earth, taking the value of 6371.004km.

In SDNE model, the encoder is from $x_i$ to $y_i^{(k)}$, the decoder is from $y_i^{(k)}$ to $\widehat{x}_i$, $y_i^{(k)}$ is the node embedding of $v_i$, in this paper,$y_i^{(k)} = p_i$. The formula of encoder is:

\begin{equation}
\boldsymbol{y}_{\boldsymbol{i}}^{(\boldsymbol{k})}=\sigma\left(\boldsymbol{W}^{(\boldsymbol{k})} \boldsymbol{y}_{\boldsymbol{i}}^{(\boldsymbol{k}-\mathbf{1})}+b^{(k)}\right)
\end{equation}

Therefore, the spatial embedding can be expressed as $\bf P$ =$\left(p_{1}, p_{2}, ..., p_{n}\right)$.
After we get the statistic embedding, the text embedding and the spatial embedding.
We combine the above features into a multi-source feature $M=(S, T, P)$, and use the fully connected neural network to predict the rental price.
We take the multi-source feature matrix $\bf M$ =$\left(m_{1}, m_{2}, ..., m_{n}\right)$ as the input of the neural network, then obtain as follows:

\begin{equation}
\boldsymbol{y}=w^T \boldsymbol{m}+b, \quad A=\sigma(\boldsymbol{y})
\end{equation}

where $y$ is the actual rental price, $m$ is the multi-source feature, $w$ is the parameter matrix, and $b$ is the offset term. 
$A$ is the activation function, we uses ReLU function as the activation function.  
At last, the output layer uses a neuron to output and get the predicted price $\widehat{y}_i$.

%% file: 6experiment.tex
\section{Experiment}

In this section, we first introduce two real dataset and evaluation metrics.
Then, we design experiments to answer the following three questions:
\begin{itemize}

\item {\bf Q1.} How is the performance of our proposed \textbf{MSIE} in the airbnb price prediction task?

\item {\bf Q2.} How do the feature combination affect the price prediction performance?

\item {\bf Q3.} What is the key influence on the airbnb price? 
\end{itemize}

\subsection{Dataset}
We collect the dataset from an open online airbnb website.
Table~\ref{table_statistic} shows the statistics of our two real airbnb datasets from two
cities: Beijing and Shanghai after preprocess.

\begin{table}[htbp] 
\caption{Statistics of the data} 
\centering
\begin{tabular}{c|c|c|c} 
			\toprule 
			 City & \# Houses & \# Reviews  & Time Period \\ 
			\midrule 
			Beijing & 10779 & 191876 & 01/2017-06/2019\\
			Shanghai & 8638 & 159069 & 01/2020-07/2021\\
			\bottomrule 
		\end{tabular} 
		\label{table_statistic}
\end{table}

Besides, we also collect the POI of the Beijing and Shanghai in Table~\ref{poi_category}.
We divide it into 8 categories, include, Education, Entertainment, Food, Beverage Shopping, Tourist, Transportation, Medical Service, and Public Service.

\begin{table}[tbp] 
\caption{POI Categories} 
\centering
\begin{tabular}{c|c|c|c} 
			\toprule 
			 Number & POI Category Name & \#Beijing & \#Shanghai\\ 
			\midrule 
			1 & Education & 8711 & 2635\\
			2 & Entertainment & 6501 & 2607\\
			3 & Food & 5744 & 6301\\
			4 & Beverage Shopping & 6601 & 5632\\
			5 & Tourist & 6713 & 4176\\
			6 & Transportation & 4322 & 1753\\
			7 & Medical Service & 3660 & 2862\\
			8 & Public Service & 5976 & 3699\\
			\bottomrule 
		\end{tabular} 
		\label{poi_category}
\end{table}

\subsection{Evaluation Metrics}
We evaluate the model performances in terms of the following metrics.

\noindent {\bf (1) $\textbf{Mean Absolute Error} (\textbf{MAE})$} represents the average of the absolute value of the error between the predicted value and the true value.
\begin{equation}
    \textbf{MAE} = \frac{1}{n}\sum_{i=1}^{n}\left | \hat{y}-y \right |
\end{equation}

\noindent {\bf (2) $\textbf{Mean Squared Error}(\textbf{MSE})$} is a measure of the closeness of the predicted value relative to the actual value.
\begin{equation}
    \textbf{MSE} = \frac{1}{n}\sum_{i=1}^{n}\left ( \hat{y}- y\right )^{2}
\end{equation}

\noindent {\bf (3) $\textbf{Root Mean Squared Error}(\textbf{RMSE})$} is defined as follows:
\begin{equation}
    \textbf{RMSE} = \sqrt{\frac{1}{n}\sum_{i=1}^{n}\left ( \hat{y}- y\right )^{2}}
\end{equation}
where $\hat{y}$ is the predicted price from the regression and $y$ is the actual price. The lower the RMSE, the better the method.

\noindent {\bf (4) $\textbf{The coefficient of determination}(R^{2})$} convert the predicted results into accuracy, the results are between [0,1].

\begin{equation}
R^2=1-\frac{\sum_{i=1}^n\left(\hat{y}_i-y_i\right)^2}{\sum_{i=1}^n\left(\bar{y}_l-y_i\right)^2}
\end{equation}

The higher the value of  $R^{2}$ , the more accurate the estimation method.

\begin{figure*}[htbp]
\setlength{\abovecaptionskip}{-0.01cm} 
\centering
\subfigure[MAE]{
\begin{minipage}[t]{0.248\linewidth}
\centering
\includegraphics[width=1\linewidth]{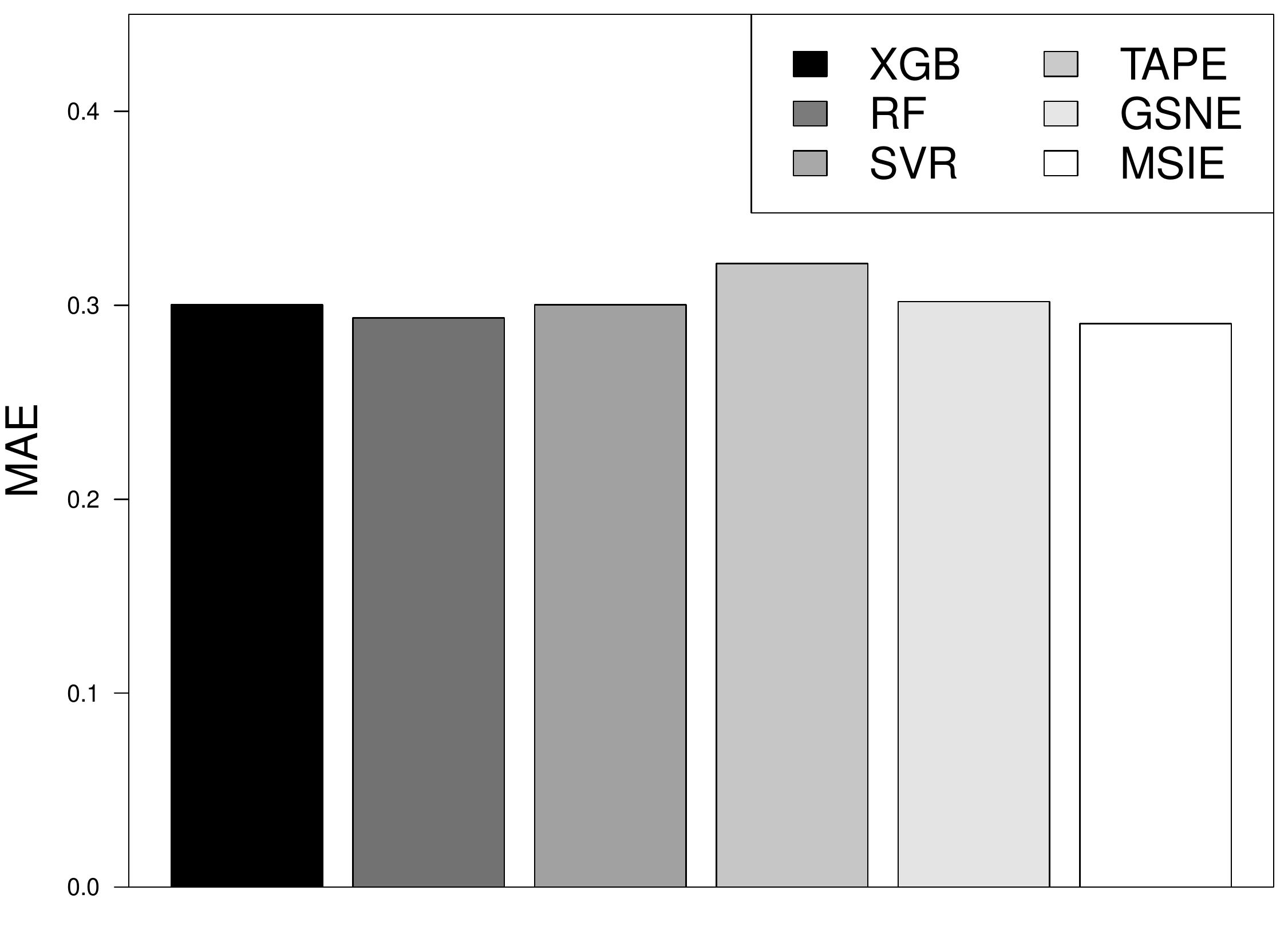}
\label{mae}
\end{minipage}%
}%
\subfigure[MSE]{
\begin{minipage}[t]{0.248\linewidth}
\centering
\includegraphics[width=1\linewidth]{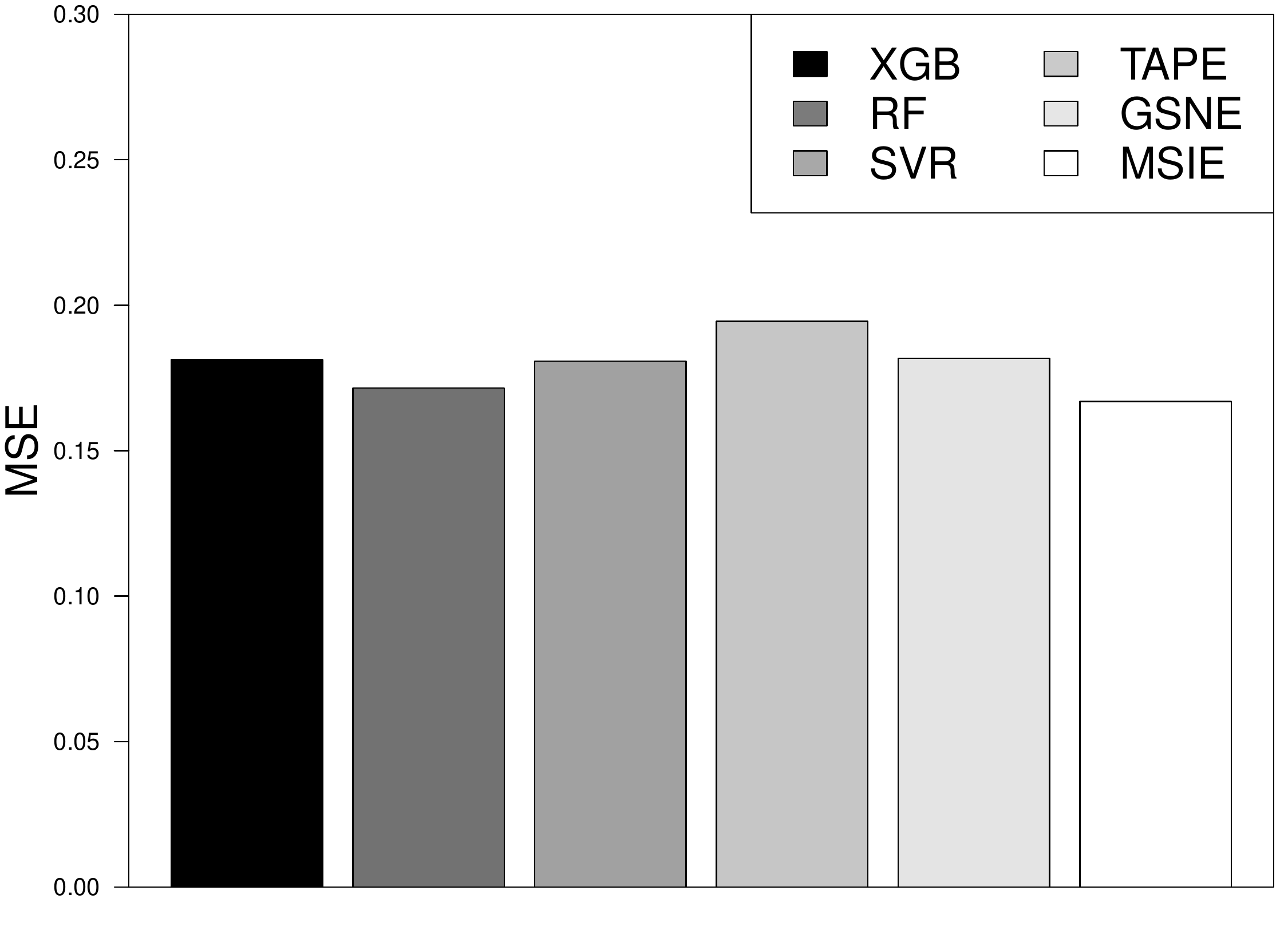}
\label{mse}
\end{minipage}%
}%
\subfigure[RMSE]{
\begin{minipage}[t]{0.248\linewidth}
\centering
\includegraphics[width=1\linewidth]{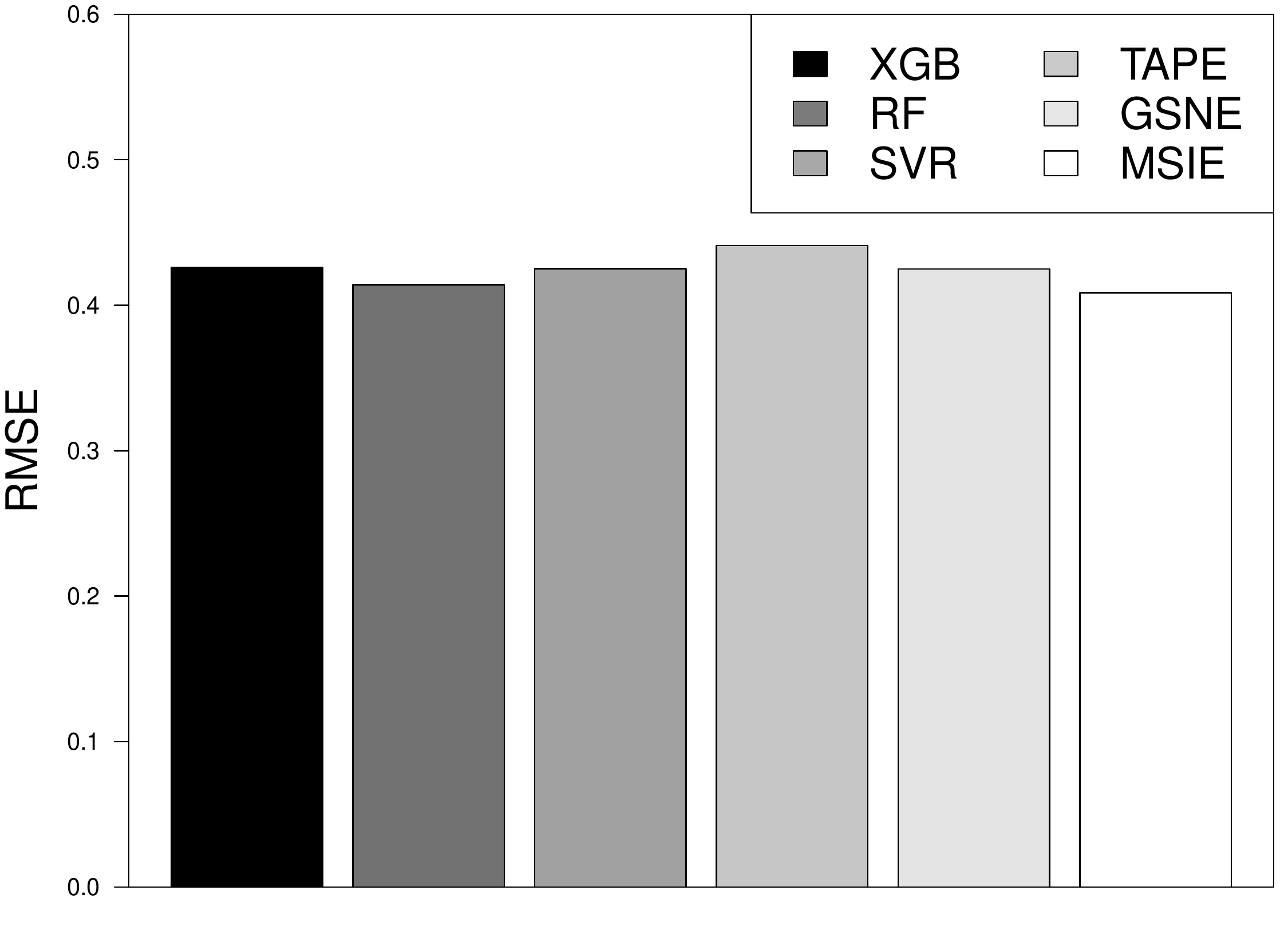}
\label{rmse}
\end{minipage}%
}%
\subfigure[$R^2$]{
\begin{minipage}[t]{0.248\linewidth}
\centering
\includegraphics[width=1\linewidth]{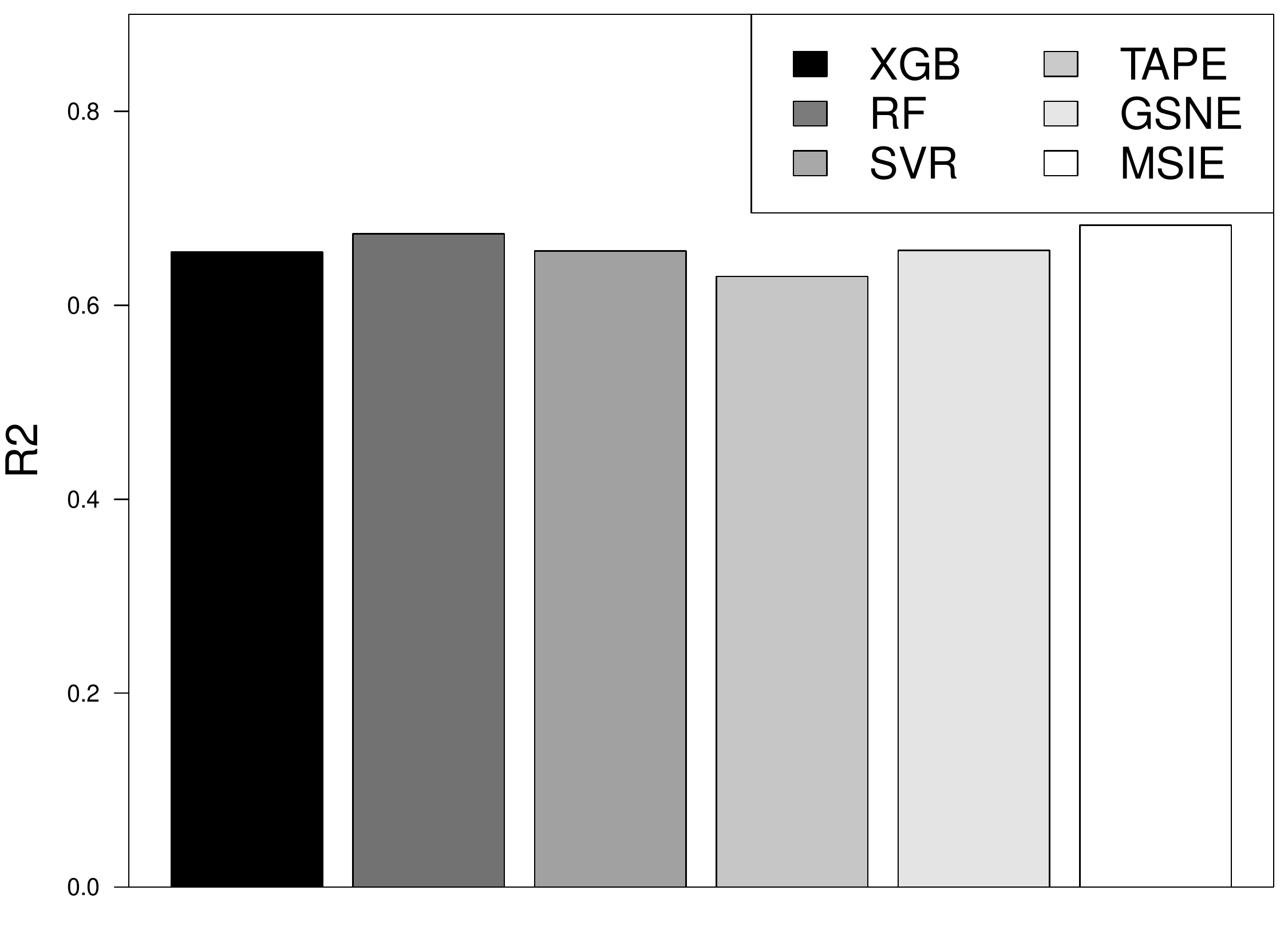}
\label{R2}
\end{minipage}
}%

\centering
\caption{Overall comparison on Beijing dataset.}
\label{figure_overall_beijing}
\end{figure*}

\begin{figure*}[htbp]
\setlength{\abovecaptionskip}{-0.01cm} 
\centering
\subfigure[MAE]{
\begin{minipage}[t]{0.248\linewidth}
\centering
\includegraphics[width=1\linewidth]{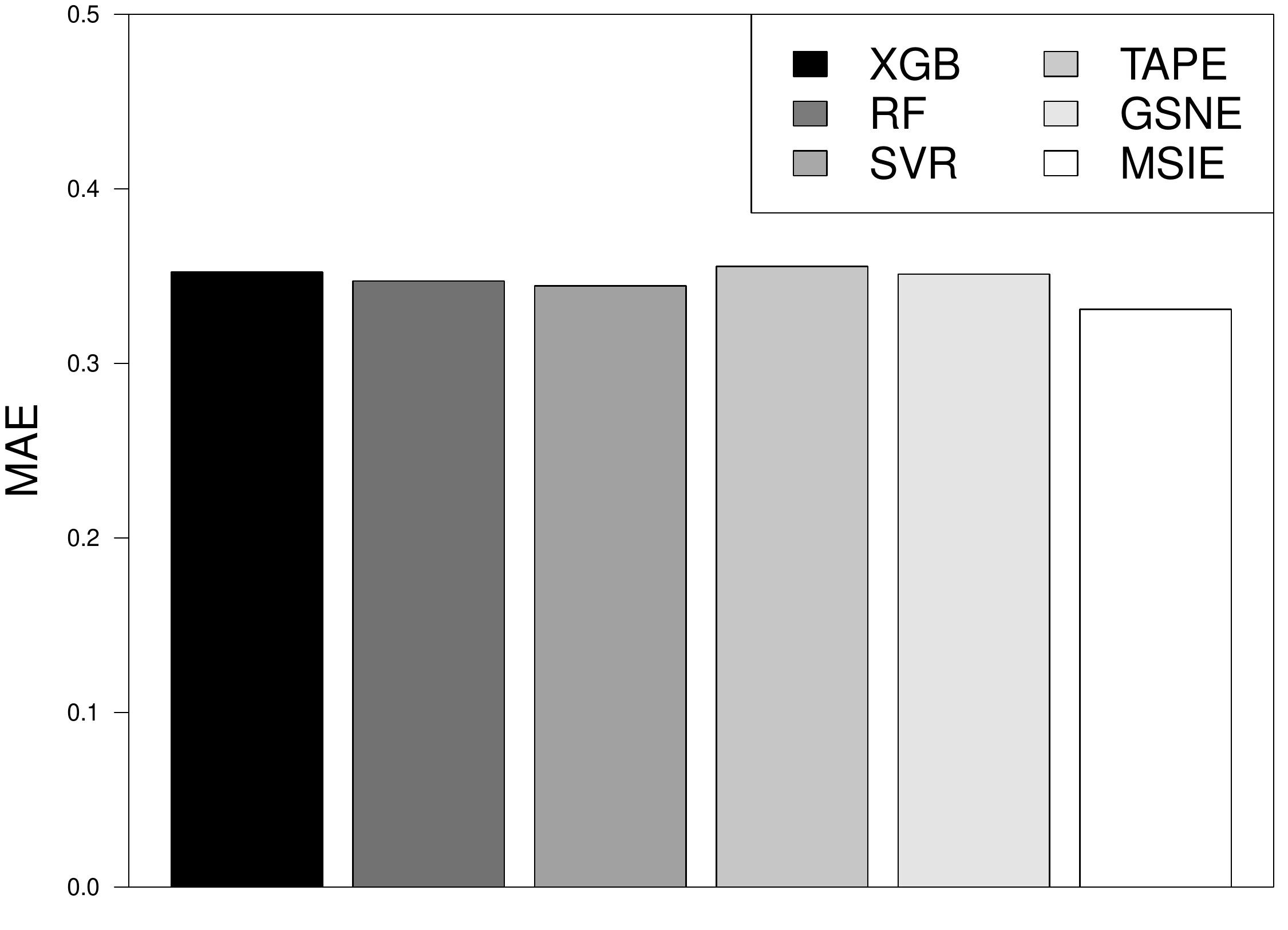}
\label{mae}
\end{minipage}%
}%
\subfigure[MSE]{
\begin{minipage}[t]{0.248\linewidth}
\centering
\includegraphics[width=1\linewidth]{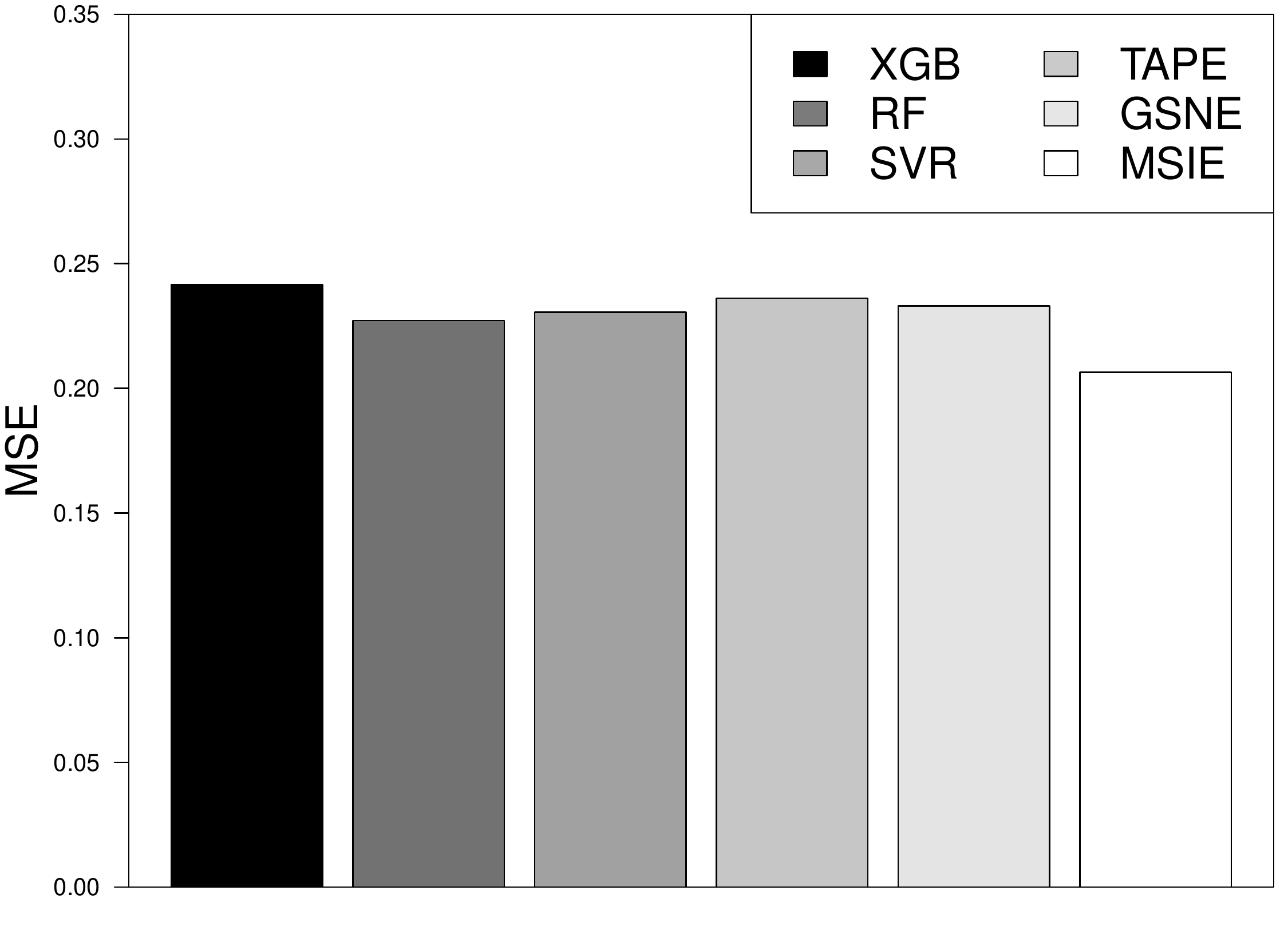}
\label{mse}
\end{minipage}%
}%
\subfigure[RMSE]{
\begin{minipage}[t]{0.248\linewidth}
\centering
\includegraphics[width=1\linewidth]{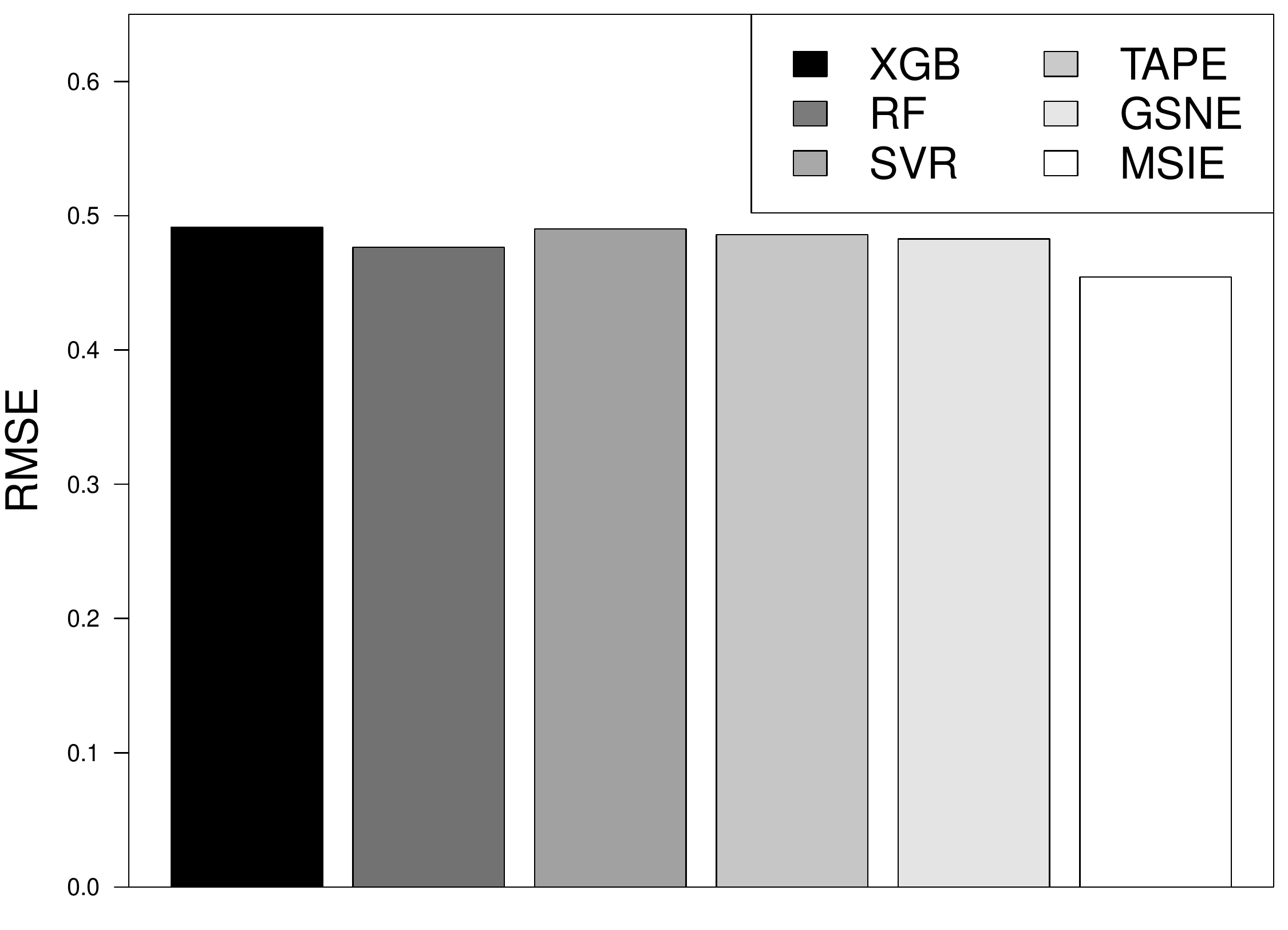}
\label{rmse}
\end{minipage}%
}%
\subfigure[$R^2$]{
\begin{minipage}[t]{0.248\linewidth}
\centering
\includegraphics[width=1\linewidth]{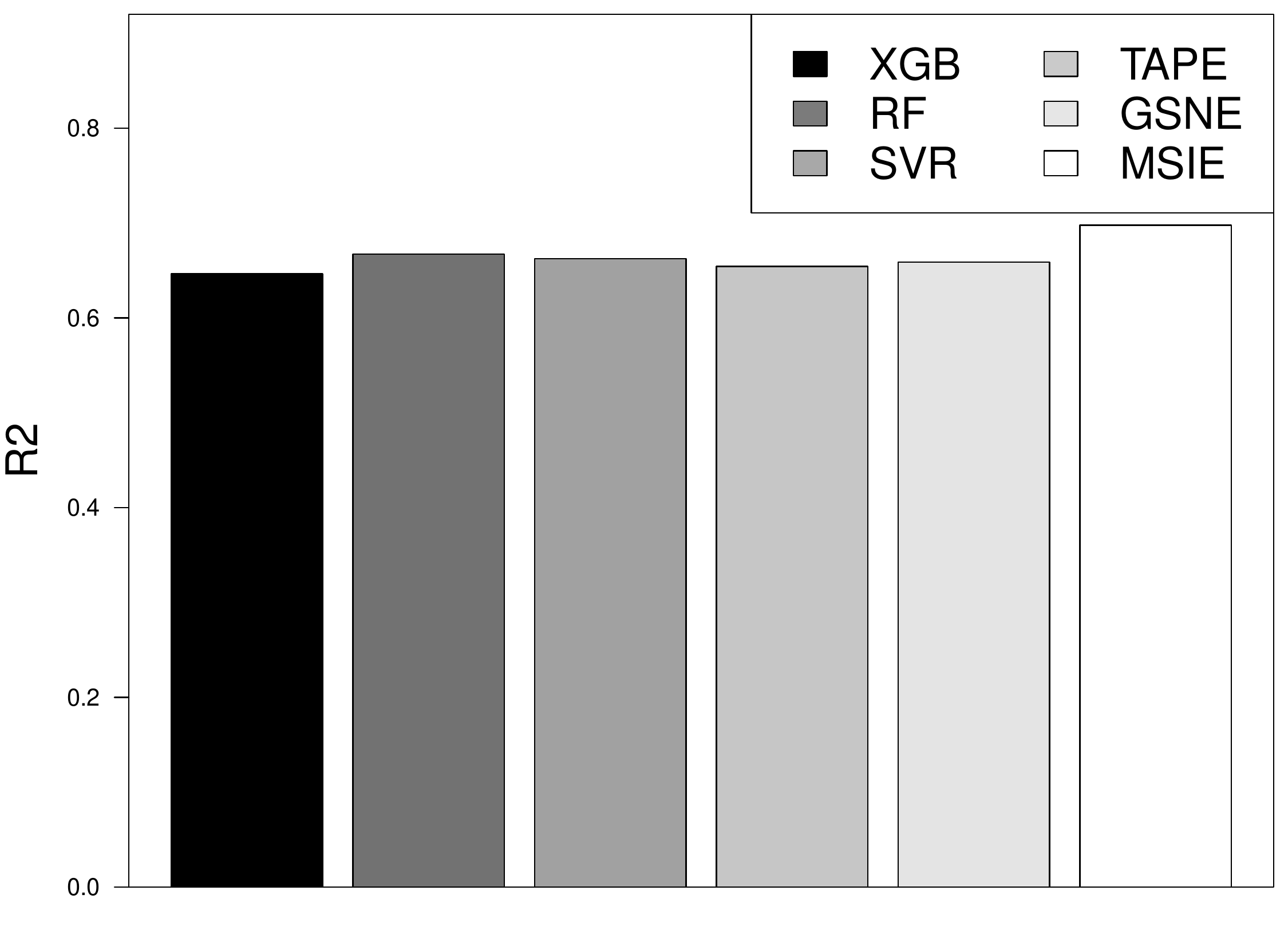}
\label{R2}
\end{minipage}
}%

\centering
\caption{Overall comparison on Shanghai dataset.}
\label{figure_overall_shanghai}
\end{figure*}

\subsection{Baseline Algorithms}
To prove the effectiveness of our model, we compare our method with the following algorithms.

\textbf{(1)Extreme Gradient Boosting}
XGBOOST~\cite{2016XGBoost} is an improvement on the boosting algorithm based on Gradient Boosting Decision Tree to make it faster and more efficient.

\textbf{(2)Random Forest}
RF~\cite{DBLP:journals/ml/Breiman01} is an algorithm that integrates multiple trees through the idea of integrated learning. Its basic unit is the decision tree.

\textbf{(3)Support Vector Regression}
SVR~\cite{Awad2015} is an algorithm that applies support vector machine to regression problems.

\textbf{(4)TAPE}
TAPE~\cite{DBLP:journals/bigdatama/ShenLCJ20} analyzed the relationship between the description of each rental and the price, and added the geographical factor component to recommend a reasonable price for each new rental of the landlord.

\textbf{(5)GSNE}
GSNE~\cite{DBLP:journals/datamine/DasALKS21} is a geospatial embedding framework, which can accurately capture the geospatial neighborhood relationship between houses and surrounding POIs. Essentially, it is to learn the low dimensional Gaussian embedding on the geospatial network node, and can be combined with the regression method, which has a certain effect on house price prediction.

Besides, our proposed model has three variants of the feature set combination:
{\bf (1) MSIE-S}, where the model utilizes the statistic feature; 
{\bf (2) MSIE-ST}, where the model utilizes the statistic feature and text feature;
{\bf (3) MSIE-STP}, where the model utilizes the statistic feature, text feature and spatial feature.
We evaluate these three variants with our model.

In the experiment, we split the dataset into two nonoverlapping sets: for all records, the earliest $80\%$ of records are the training set and the remaining $20\%$ are testing set. 
We implement the model by Pytorch and run the code
on Windows10, Inter(R) Core(TM) i7-7700HQ @2.80GHZ and memory size 8G.

\subsection{Overall Performances}

We present the results for “MAE”, “MSE”, “RMSE” and “$R^2$”, compared with baseline algorithms.
Figure~\ref{figure_overall_beijing} and Figure~\ref{figure_overall_shanghai} show that our proposed method "MSIE" outperform the baselines over both the Beijing and Shanghai dataset.
The lower value of "MAE", "MSE", "RMSE", and the higher value of ”$R^2$”, means the performance better.
In all cases, we observe an improvement with respect to baseline algorithms, especially on “MSE” and “$R^2$”.
One interesting observation is that the traditional machine learning method(such as, "XGB", "RF" and "SVR") performs better than "TAPE".  
We analysis the reason is that our algorithm feature engineering proposed in this paper is well done and has universality.

\begin{figure}[tbp]
\centering
\subfigure[Beijing]{
\begin{minipage}[t]{0.49\linewidth}
\centering
\includegraphics[width=1\linewidth]{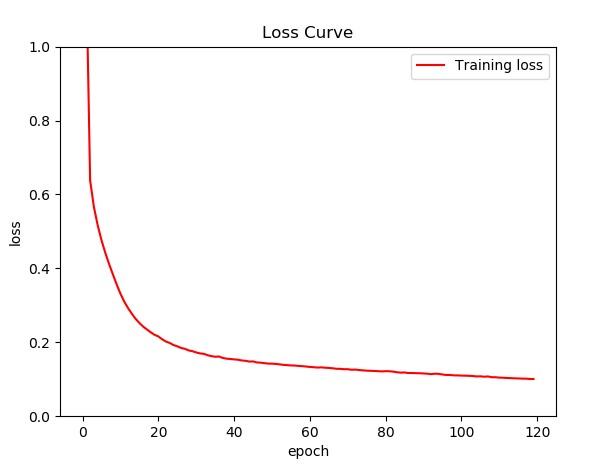}
\label{mae}
\end{minipage}%
}%
\subfigure[Shanghai]{
\begin{minipage}[t]{0.48\linewidth}
\centering
\includegraphics[width=1\linewidth]{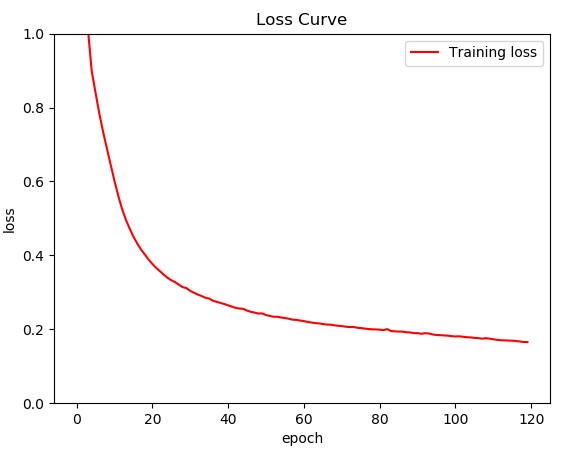}
\label{mse}
\end{minipage}%
}%
\centering
\caption{The loss curve on two datasets.}
\label{figure_loss}
\end{figure}

Besides, we uses the fully connected neural network constructed as the prediction model. 
In order to prevent over fitting, we sets 128 neurons in the input layer, uses 2 hidden layers and set Epoch=120, the size of Batch as 256. 
Figure~\ref{figure_loss} show the loss curves of neural network training on the two datasets respectively.

\subsection{Robustness Check}

We evaluate the feature embedding contribution on modeling representations.
To set the control group, we develop a variant of the proposed "MSIE", namely "MSIE-S", "MSIE-ST", "MSIE-STP".
"MSIE-S", "MSIE-ST", "MSIE-STP" takes the different combination of feature set as the input, while other component of remains the same.
Table~\ref{table_featureset_beijing} and Table~\ref{table_featureset_shanghai} show the comparison results.
We can observe that the performance of "MSIE-STP" outperforms "MSIE-S" and "MSIE-ST" in terms of the four metrics over both two datasets.
The results validate that the integration of text feature and spatial feature indeed enhances the modeling of price prediction. 

\begin{table}[tbp]
\caption{The feature combination on Beijing dataset.}
\begin{center}
\begin{tabular}{p{60pt}|p{30pt}|p{30pt}|p{30pt}|p{30pt}}
\hline
\textbf{Feature set}&\textbf{MAE}&\textbf{MSE}&\textbf{RMSE}&\textbf{$R^2$}\\
\hline
MSIE-S & 0.3652 & 0.2341 & 0.4839 & 0.5545\\
\hline
MSIE-ST & 0.2941 & 0.1688 & 0.4109 & 0.6786\\
\hline
MSIE-STP & 0.2905 & 0.1669 & 0.4086 & 0.6824\\
\hline
\end{tabular}
\label{table_featureset_beijing}
\end{center}
\end{table}

\begin{table}[tbp]
\caption{The feature combination on Shanghai dataset.}
\begin{center}
\begin{tabular}{p{60pt}|p{30pt}|p{30pt}|p{30pt}|p{30pt}}
\hline
\textbf{Feature set}&\textbf{MAE}&\textbf{MSE}&\textbf{RMSE}&\textbf{$R^2$}\\
\hline
MSIE-S & 0.4003 & 0.2852 & 0.5340 & 0.5824\\
\hline
MSIE-ST & 0.3512 & 0.2371 & 0.4869 & 0.6527\\
\hline
MSIE-STP & 0.3310 & 0.2065 & 0.4544 & 0.6977\\
\hline
\end{tabular}
\label{table_featureset_shanghai}
\end{center}
\end{table}

\subsection{Analysis of Key Influence}

\begin{figure}[htbp]
\centering
\subfigure[Beijing]{
\begin{minipage}[t]{0.48\linewidth}
\centering
\includegraphics[width=1\linewidth]{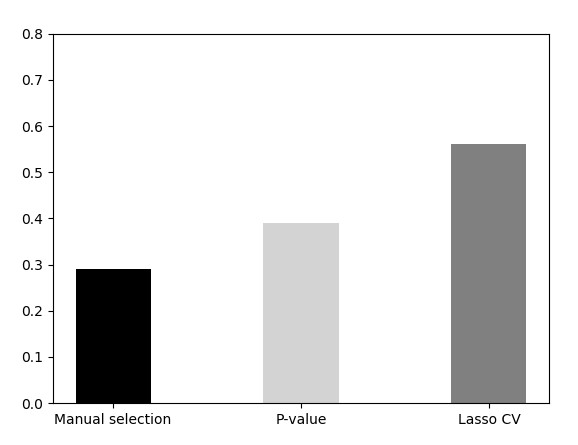}
\label{mae}
\end{minipage}%
}%
\subfigure[Shanghai]{
\begin{minipage}[t]{0.49\linewidth}
\centering
\includegraphics[width=1\linewidth]{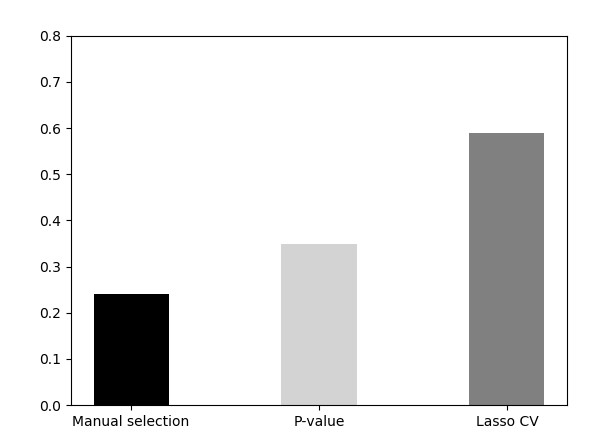}
\label{mse}
\end{minipage}%
}%
\centering
\caption{The $R^2$ with different feature selection method.}
\label{figure_lassocv}
\end{figure}

In order to analysis the key feature-influence of price, according to the previous studies, we use three feature selection method, include manual selection, P-value~\cite{pvalue02} and Lasso CV~\cite{doi:10.1198/016214506000000735}.
We use $R^2$ as an indicator to analyze, and the results are shown in Figure~\ref{figure_lassocv}. 
The best result is to use Lasso CV to select features from the original data.

\begin{figure}[htbp]
\centering
\subfigure[Beijing]{
\begin{minipage}[t]{0.485\linewidth}
\centering
\includegraphics[width=1\linewidth]{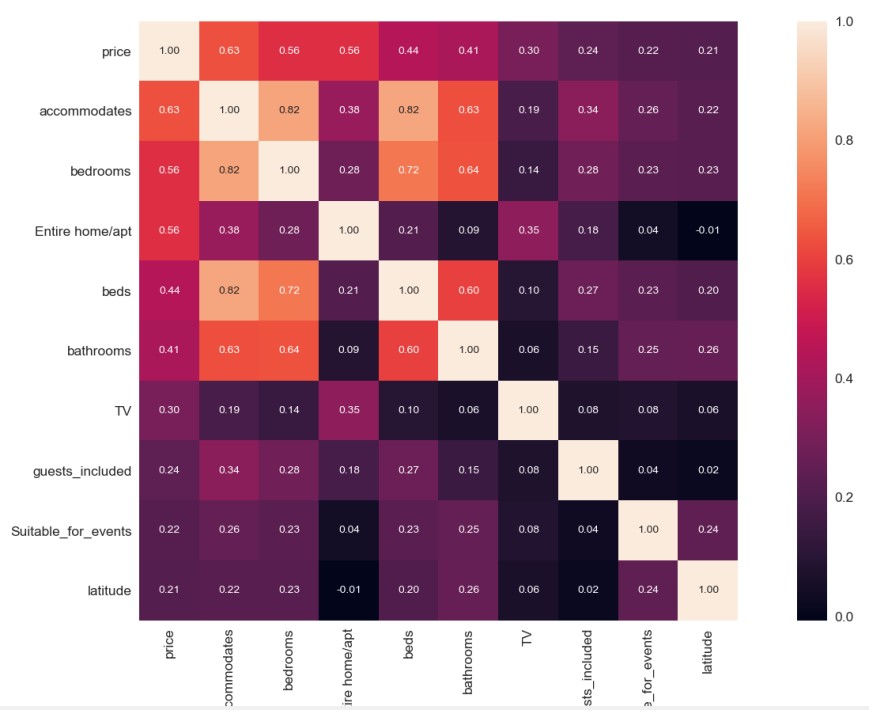}
\label{mae}
\end{minipage}%
}%
\subfigure[Shanghai]{
\begin{minipage}[t]{0.485\linewidth}
\centering
\includegraphics[width=1\linewidth]{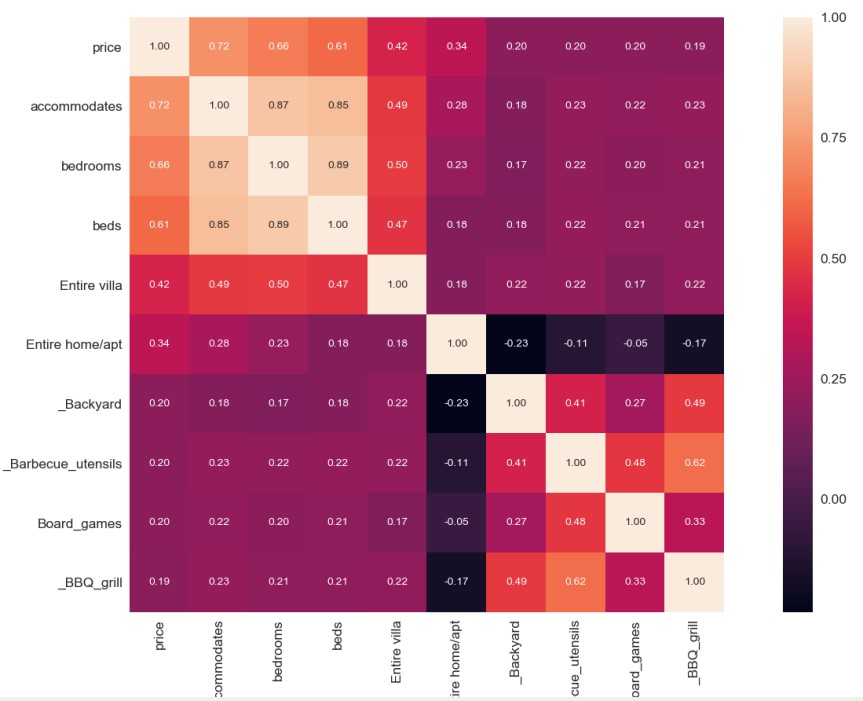}
\label{mse}
\end{minipage}%
}%
\centering
\caption{The feature most relevant to price.}
\label{figure_pvalue}
\end{figure}

Then, we selects 20 features with the highest correlation for rental price prediction according to P-value, and uses Lasso CV to select features to obtain statistical information as input for prediction. 
Figure~\ref{figure_pvalue} show the 10 features with the highest correlation with rental prices in the two datasets.

%% file: 7relatedwork.tex
\section{Related work}
In this work, we propose a model to predict the price of listings on Airbnb. 
Some studies used sentiment analysis to study the problem of Airbnb. 
Martinez R D {\it et al.} studied the relationship between Airbnb host's listing description and occupancy rates by sentiment analysis~\cite{2017The}. 
Zhang {\it et al.} proposed a text analytics framework to study the relationship among self-description, trust perception and purchase behavior on Airbnb. 
They used text mining to extracted sentiment intensity and use regression method to identify the impact of linguistic and semantic features on trust perception~\cite{2020A}. 
Kalehbast P R {\it et al.} used sentiment analysis and machine learning to predict the price of listings on Airbnb~\cite{2019Airbnb}.

Most researches have proposed many works to study the price by using the listing information.
Wang {\it et al.} studied the relationship between a price and its determinants by various listing information (e.g., host identity verified, accommodates, wireless Internet, amenities and services, and free parking)~\cite{2017Price}. 
P.Choudhary {\it et al.} analyzed Airbnb listings in San Francisco to better understand how different attributes (e.g., bedrooms, location, and listing type) can be used to accurately predict the price of a new listing, which is optimal in terms of the host’s profitability yet affordable to their guests~\cite{2018Unravelling}.
Shen {\it et al.} analyzed the relationship between the description of each listing and its price, and proposed a text-based model to recommend a reasonable price for newly added listings~\cite{2020Text}. 
Tang {\it et al.} labeled text information for nine handpicked classes, extracted image-related features, and finally used all features to predict a listing’s neighborhood and its price~\cite{0Neighborhood}.

%% file: 8conclusion.tex
\section{Conclusion}

In this paper, we proposes a prediction model based on multi-source information embedding to study the Airbnb price problem. 
Specifically, in order to obtain the best feature set, we first selects the features of the house itself to obtain statistical information features.
Secondly, the text information in this paper is divided into three categories, and the house description and landlord introduction are converted into feature matrix.
The tenant reviews are then converted into sentiment scores about each house. 
Then, we uses different types of point-of-interest (POI) data and houses to form various spatial network graphs and learns their network embeddings to obtain spatial information features.  
Finally, we combines these three types of feature embeddings are combined into multi-resource housing features as input, and the neural network constructed in this paper is used for price prediction. The effectiveness of our model is demonstrated with two real data.
For future work, we plan to combine some heuristic
methods~\cite{DBLP:journals/ai/WangCCY20,pan2023improved} to further improve performance.



\section*{Acknowledgments}

This work is supported by the Natural Science Research Foundation of Jilin Province of China under Grant No. YDZJ202201ZYTS415, the Fundamental Research Funds for the Central Universities 2412019ZD013, NSFC (under Grant Nos.61976050 and 61972384).